\documentclass[twocolumn,final]{elsarticle}

\usepackage{medima}
\usepackage{framed,multirow}

\usepackage{amssymb}
\usepackage{amsmath}
\usepackage{physics}
\usepackage{booktabs}
\usepackage[ruled,vlined]{algorithm2e}
\usepackage{url}
\usepackage{latexsym}
\usepackage{soul}
\usepackage{xspace}
\usepackage{subcaption}
\usepackage{hyperref}

\usepackage[capitalize]{cleveref}
\crefname{section}{Sec.}{Secs.}
\Crefname{section}{Section}{Sections}
\Crefname{table}{Table}{Tables}
\crefname{table}{Tab.}{Tabs.}
\definecolor{newcolor}{rgb}{.8,.349,.1}

\journal{Medical Image Analysis}

\def\myarch{RadFormer\xspace}

\newcommand{\beginsupplement}{%
    \setcounter{table}{0}
    \renewcommand{\thetable}{S\arabic{table}}%
    \setcounter{figure}{0}
    \renewcommand{\thefigure}{S\arabic{figure}}%
 }

\begin{document}

\verso{Soumen Basu \textit{et~al.}}

\begin{frontmatter}

\title{\myarch: Transformers with Global-Local Attention for Interpretable and Accurate Gallbladder Cancer Detection}


\author[1]{Soumen \snm{Basu}\corref{cor1}}
\cortext[cor1]{Corresponding author.}
\ead{soumen.basu@cse.iitd.ac.in}
\author[1]{Mayank \snm{Gupta}}
\author[2]{Pratyaksha \snm{Rana}}
\author[2]{Pankaj \snm{Gupta}}
\author[1]{Chetan \snm{Arora}}

\address[1]{Department of Computer Science, Indian Institute of Technology Delhi, New Delhi, India}
\address[2]{Department of Radiodiagnosis and Imaging, Postgraduate Institute of Medical Education \& Research, Chandigarh, India}

\received{}
\finalform{}
\accepted{}
\availableonline{}
\communicated{}

\begin{abstract}
	We propose a novel deep neural network architecture to learn interpretable representation for medical image analysis. Our architecture generates a global attention for region of interest, and then learns bag of words style deep feature embeddings with local attention. The global, and local feature maps are combined using a contemporary transformer architecture for highly accurate Gallbladder Cancer (GBC) detection from Ultrasound (USG) images. Our experiments indicate that the detection accuracy of our model beats even human radiologists, and advocates its use as the second reader for GBC diagnosis. Bag of words embeddings allow our model to be probed for generating interpretable explanations for GBC detection consistent with the ones reported in medical literature. We show that the proposed model not only helps understand decisions of neural network models but also aids in discovery of new visual features relevant to the diagnosis of GBC. Source-code is available at \url{https://github.com/sbasu276/RadFormer}
\end{abstract}


\end{frontmatter}

\section{Introduction}
Recently, Deep Neural Networks (DNNs) have shown remarkable performance in a variety of medical imaging tasks, including cancer detection \citep{kooi2017large, codella2017deep, langlotz2019roadmap, chu2019application, pirovano2021computer}, polyp segmentation \citep{wu2021collaborative, wu2021precise, wu2022polypseg}, or ventricular segmentation \citep{wu2021automated, wu2022semi}. Using DNNs to assist medical experts in diagnosis and medical procedures is a rapidly growing area of research. However, although DNNs have achieved super-human accuracy in a plethora of medical imaging tasks, the black-box nature severely limits their usage in a practical clinical setting. 

\par A diagnostic system must provide adequate explanations for the diagnosis, transparently and comprehensively, to earn the trust of medical experts and patients. Further, there are regulations and laws for using machine learning systems in clinical settings. For example, European General Data Protection Regulation (GDPR) mandates healthcare organizations to provide on-demand explanations of diagnostic decisions \citep{hoofnagle2019european}. Thus, it is critical to integrate the ability to provide interpretation and explanation of its decisions in machine learning models. On the other hand, the success of DNNs is primarily attributed to the enormous parametric space and robust learning algorithms. Millions of parameters and complex dependencies between the activations makes it extremely challenging to interpret DNN predictions.  

\par Explainable Artificial Intelligence (XAI) aims to design systems that allow human users to comprehend how the AI systems such as DNNs reach a particular decision. Even though there has been prior work on the interpretability of natural images \citep{bau2017network, monga2021algorithm}, developing XAI systems for specialized medical applications such as cancer detection remains a challenge. Previously, DNN models have been developed for diseases like breast cancer \citep{bejnordi2017diagnostic, lamy2019explainable}, lung cancer \citep{coudray2018classification}, or brain cancer \citep{pg-cam, ismael2020enhanced}. However, the explainability of these models predominantly depends on heatmap visualization, which identifies the salient regions for the general audience but lacks specific medical concepts for interpretability. Expert radiologists use certain features defined in reporting standards to characterize a pathology. For example, loss of interface with liver or extramural invasion are features of a malignant gallbladder \citep{gb-rads-paper}. We denote such features used by medical experts as \emph{radiological features} or \emph{lexicons}. In contrast to the heatmap visualizations, we use the visual bag-of-words style feature embedding of the local region of interest for providing precise explanations of decisions. Such visual bag-of-features are easy to map to radiological features of the pathologies, and are consistent with the reporting standards while providing explanations of network decisions. We do not use the radiological feature annotations during the training of our architecture. The global image-level labels suffice. Towards this end, we explore the interpretability of DNNs for Gallbladder Cancer (GBC) detection on Ultrasound (USG) images. We incorporate the specific medical concepts to improve the explainability of decisions. Despite the applicability of machine learning in other types of cancer detection, limited study on the usage of DNNs for detecting Gallbladder Cancer (GBC) on USG exist.

\par GBC is the most common biliary tract malignancy and the fifth most common malignancy of the gastrointestinal tract. According to GLOBOCAN 2018, GBC causes 165,087 deaths and 219,420 incidences worldwide, every year \citep{bray2018global}. GBC is among the very few types of cancers which manifest a higher proportion of mortality than incidence. GBC is overly deadly because it is rarely detected before an advanced and metastasized stage in most patients, impeding curative resection and resulting in a dismal prognosis \citep{batra2005gallbladder, randi2006gallbladder}. Although GBC is relatively rare, its high mortality rate signifies the criticality of early diagnosis to improve the survival statistics of the deadly disease. 

\par Ultrasound  (USG) is most widely used diagnostic imaging modality due to its non-ionizing radiation, low cost, and accessibility \citep{klibanov2015ultrasound}. Ultrasound is the first-line imaging test for evaluation of gallbladder diseases. It is often the sole diagnostic test performed for patients with suspected Gallbladder (GB) diseases. No further testing is usually performed if malignancy is not detected in the preliminary GB screening and GBC progresses silently. Identifying stones or GB wall thickening at routine USG is easy, but the precise characterization of GBC is challenging \citep{gupta2020imaging}. Furthermore, applying DNNs to USG image analysis poses major hurdles. USG images, unlike MRI or CT, contain significant noise and artifacts. The hand-held nature of the ultrasound sensors adds operator specific visual variations. The artifacts like shadows often exhibit similar visual traits of a non-malignant GB region and bias the state-of-the-art DNN classifiers, leading to poor detection accuracy \citep{basu2022surpassing}. 

\par In this work, we propose, \myarch (Radiology Transformer), for detecting GBC from USG images as well as provide semantically relevant explanation of the diagnosis like a radiologist. The key contributions of our work are following:
\begin{enumerate}
	\item We develop \myarch, which uses a transformer-based architecture to efficiently integrate the global-local attention and produce highly accurate classification of GBC from USG images. Our model outperforms all state of the art DNN models for the GBC detection tasks: accuracy 0.92 by our model compared to 0.84 by the state of the art. It even beats human radiologists in terms of accuracy of detection in our experiments: 0.70 and 0.68 by the two expert radiologists.
	\item Our design of \myarch exploits the local bag-of-features for extracting the radiological features consistent with the reporting standards to provide detailed explanation of the GBC diagnosis. \myarch does not require any special annotation and uses only global image level labels for training the explanation module.
\end{enumerate}

\section{Related Work}
\subsection{Explainability in DNNs for cancer detection}
Explainability of a DNN prediction has been explored for various kinds of cancer detection problems such as breast, liver, brain, or lung. Breast cancer in particular has received significant attention from the community \citep{dhungel2016automated, zhou2018radiomics, samala2018breast}. While \cite{wang2018breast} use the VGGNet and Grad-CAM to visualize salient regions of breast cancer in mammography, \cite{wu2018deepminer} attempt to learn representations which help explainability. A study by \cite{hamm2019deep} uses last layer feature maps to identify attention regions while classifying liver lesions in multiphasic MR images.  \cite{windisch2020implementation} aim to train a DNN with explainable features on MRI slices for detecting glioblastoma and vestibular schwannoma. \cite{pg-cam} show that pyramid gradient-based class activation mapping (PG-CAM) provides better localization of meningioma in MR images. \cite{venugopal2020unboxing} uses heatmaps to characterize lung nodules and assess risk of malignancy. \cite{kumar2019sisc} automatically discovers radiomics features of lung cancer through radiomics sequencing. \cite{gulum2021review} presents a comprehensive review of explainable deep models for cancer detection.
\par As discussed in the introduction, the explainability of such models, however, predominantly depends on heatmap visualization of the salient regions. Such visualization-based methods do not integrate the specific medical concepts used by medical experts in diagnosis.
\par There are two primary differences of our work with the existing methods. (1) We use a global-local neural architecture to effectively use the global contexts and the local details for better GBC detection. (2) We use a visual bag-of-words style feature embedding of the local details for providing explanations of decisions. We algorithmically map the features from the bag-of-features with radiological lexicons, and provide explanations of network decisions that are consistent with medical reporting standards.

\subsection{DNNs for gallbladder afflictions}
Multiple studies have shown the application of DNNs in detecting gallbladder ailments, such as calculi, cholecystitis, and polyps from diagnostic images. \cite{gbYolo} use YOLOv3 to detect the gallbladder and gallstones from CT images. \cite{gbPolyp} segment the gallbladder region and use AdaBoost classifier thereafter for diagnosing polyps. \cite{gbPolyp2}, on the other hand, classify neoplastic polyps from cropped samples of gallbladder USG images using an InceptionV3 model. \cite{jang2021diagnostic} used ResNet50 to diagnose polypoidal lesions on endoscopic ultrasound.
\par Although, there are several studies involving the usage of DNNs for gallbladder-related diseases, there is only a handful of studies that explore AI-based GBC detection. The study by \cite{xue2021segnet} uses a Segnet model to improve the diagnostic coincidence rate of gallbladder stones with GBC and the relationship of P16 tumor suppressor with GBC on contrast enhanced ultrasound. \cite{chang2022ct} use a UNet based denoising to improve the image quality of Low-Dose CT scans for characterizing GBC. Both of these studies primarily use the segmentation methods. \cite{basu2022surpassing} on the other hand use a specialized CNN architecture and a Gaussian blurring-based curriculum for efficient GBC detection from ultrasound images. However, none of these studies explore the interpretability.  
Further, we are not aware of any existing work on developing explainable models for GBC or gallbladder related afflictions. This motivates the current work.

\subsection{Transformers and global-local methods}
Transformer architecture was first proposed \citep{vaswani2017attention} for natural language processing tasks. Transformers are gaining popularity in the vision community due to their superior ability to model long-range dependencies \citep{dosovitskiy2020image, touvron2021training, shao2021temporal}. 
\par Transformer-based architectures are also being adopted  by the community for medical image analysis tasks. Medical transformer \citep{valanarasu2021medical} uses gated axial attention for brain anatomy segmentation in USG images. Swin-UNet \citep{cao2021swin} uses a UNet-shaped transformer architecture for segmenting abdominal organs in CT images and anatomical structures in cardiac MRI. Both of these architectures use transformers as an end-to-end network. \cite{wang2022transformer} proposed a transformer-based backbone in a contrastive learning setup. Polyp-PVT \citep{dong2021polyp} tries to segment polyps in endoscopic images using a pyramid transformer. \cite{he2021global} proposed a global-local transformer architecture to estimate brain age from MR images. 
Using global context and local information in neural architectures are also gaining popularity in medical imaging tasks \citep{guo2019deep, van2021hooknet, wu2021region, wu2022cross}. SCS-Net \citep{wu2021scs} provides a context-aware multi-scale solution to retinal vessel segmentation. \cite{wu2020automated} propose an adaptive dual attention mechanism for segmenting skin lesions from dermoscopic images.
Another prominent line of work is using a fusion of CNN-based and transformer-based features. \citep{chen2021transunet, wang2021transbts}. \cite{wu2022fat} combine both transformer and CNN-based encoders for skin lesion segmentation. Our work, on the other hand, uses a transformer for fusing global and local features coming from two different CNN backbones. The transformer in our design is specifically used as the feature fusion module.
\par We use the attention mechanism of a transformer \citep{vaswani2017attention} to optimally integrate the global and local information. The key idea of a transformer is to use a self-attention mechanism on the input sequence to capture the attention on the local patches. The input sequence is used to generate three sequences - `query', `key', and `value'. Finally, the attention is obtained using the `query' and `key' and is applied to the `value' to get an output sequence. Such an attention-base fusion of the global context and local bag-of-features boost the GBC detection performance.
\par While there are some works that use transformers for combining different kinds of features \citep{xia2021effective, he2021global}, the transformer-based feature fusion module in our design allows the local branch to generate an interpretable bag-of-feature style embedding. While both global-local architectures and transformers are generic concepts, we have utilized these concepts in order to provide semantically meaningful medical image predictions. 

\begin{figure}
    \centering
    \includegraphics[width=\linewidth]{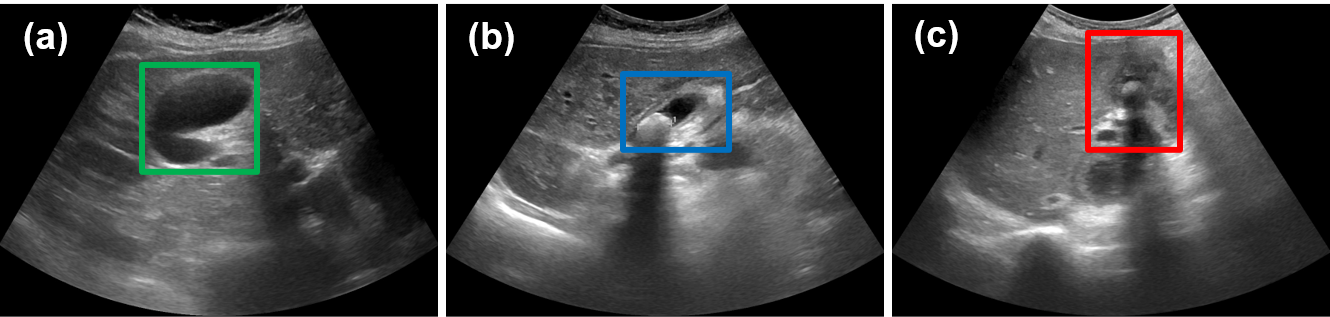}
    \caption{Sample ROI annotation. (a) Normal gallbladder with ROI in green, (b) benign GB with ROI in blue, and (c) malignant GB with ROI in red. }
    \label{fig:data_sample}
\end{figure}
%
\section{Dataset}
We have used the Gallbladder Cancer Ultrasound (GBCU) dataset \citep{basu2022surpassing}, publicly released by us in our previous work. For completeness, we provide a comprehensive description of the dataset in this section.

\subsection{Data acquisition}
The patient data samples were acquired from the radiology department of the Postgraduate Institute of Medical Education \& Research (PGIMER, India). The patients were advised 8--8 hours of fasting before the examination for the adequate distension of the gallbladder. A Logic S8 machine (GE Healthcare) with a convex low-frequency transducer with a frequency of 1--5 MHz was used to capture the USG images. Radiologists having expertise in abdominal radiology assessed the patients from various positions and angles using subcostal and intercostal views for adequate visualization of the entire gallbladder, including its fundus, neck, and body. 
A minimum of 10 Grayscale B-mode static images, including both sagittal and axial sections, were recorded for each patient.
Colour Doppler, spectral Doppler, annotations, and measurements from the dataset are excluded. All personal textual information was removed from the images to anonymize them. 

\subsection{Ground-truth labeling}
The biopsy reports of the patients are used to label each USG image as one of the three classes - normal, benign, or malignant. \myarch uses only these image-level labels during training. Apart from the image-level class labels, the bounding box annotations to locate the gallbladder and the surrounding ROI are also captured. It is important to note that we do not use these ROI bounding boxes during the training or inference of \myarch. 
Two radiologists with 8 and 3 years of experience in abdominal radiology annotated the ROIs with consensus by using a single free-size axis-aligned box spanning the gallbladder and the adjacent regions. \cref{fig:data_sample} shows sample ROI annotations.

\subsection{Statistics}
From the collected image corpus, the radiologists labeled 1255 USG images acquired from 218 patients. Our dataset contains 71 patients with a normal gallbladder, 100 patients with benign abnormality, and 47 patients with malignancy. Overall, the dataset contains 432 normal, 558 benign, and 265 malignant images. The images have a width of 801--1556 pixels and a height of 564--947 pixels. The variable size is due to cropping out of patient information from the image.

\subsection{Dataset splits}
Given the relatively small dataset, we report 10-fold cross-validation metrics for various experimental models on the entire dataset. All images of any particular patient appeared either in the training or the validation split during the cross-validation to guarantee the generalization to unseen patients. Additionally we discuss the comparison with human experts using the GBCU test split containing 122 images (31 normal, 49 benign, and 42 malignant).

\begin{figure*}[t]
    \centering
    \includegraphics[width=\textwidth]{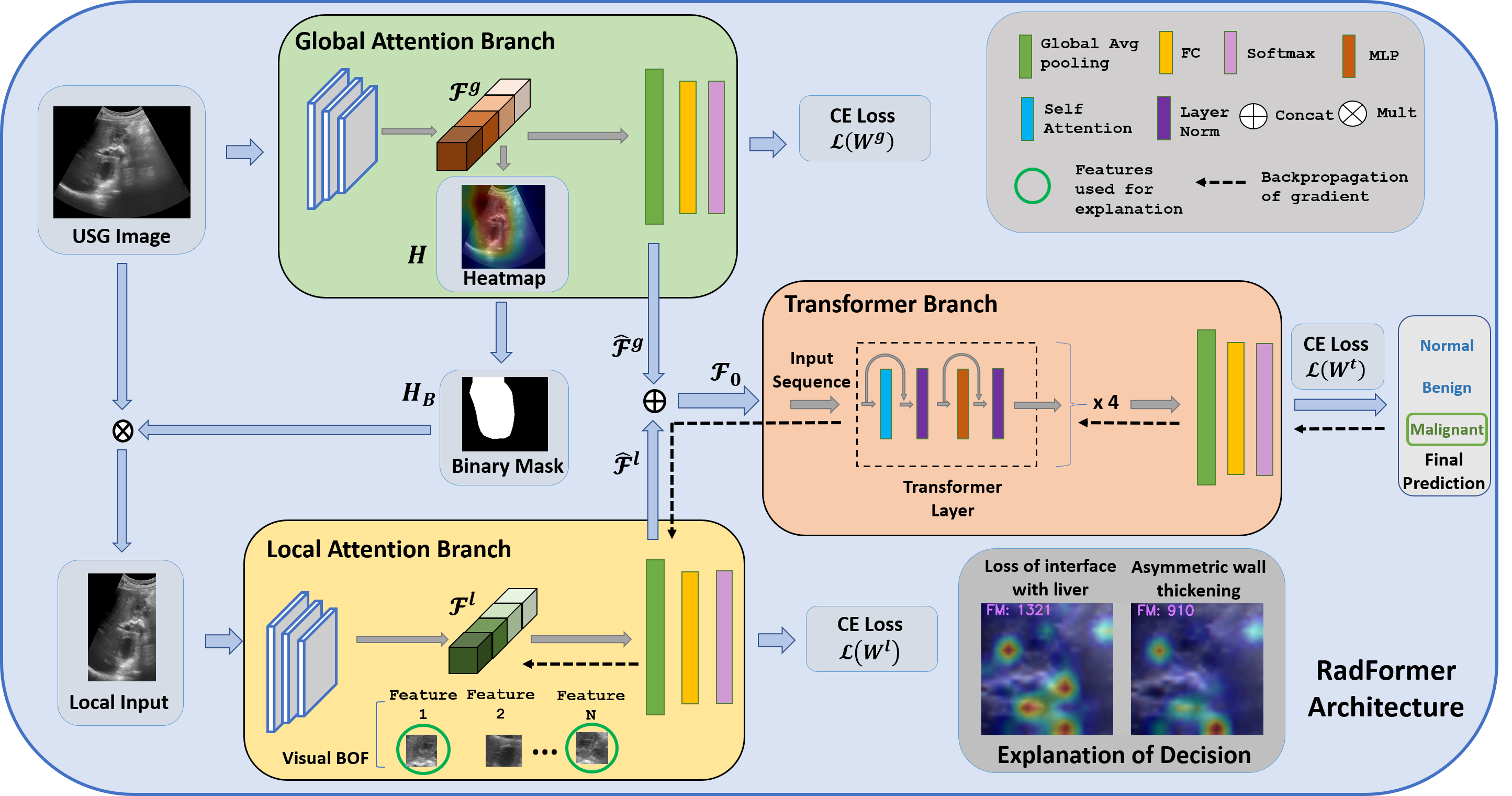}
    \caption{A schematic diagram of the \myarch architecture. The global attention branch localizes the region of interest (ROI) featuring the gallbladder. The local attention branch uses a visual bag of words style deep feature embedding for radiology-standard interpretations. The transformer efficiently fuse the global and local features for a superior performance. We note that, backpropagation of gradients is applied to both global and local branches during the training. However, as the local branch generates the bag-of-features, we use the backpropagation on local branch for retrieving the visual features relevant for interpreting the diagnosis at inference time. }
    \label{fig:arch}
\end{figure*}
%


\begin{figure}[t]
    \centering
    \includegraphics[width=\linewidth]{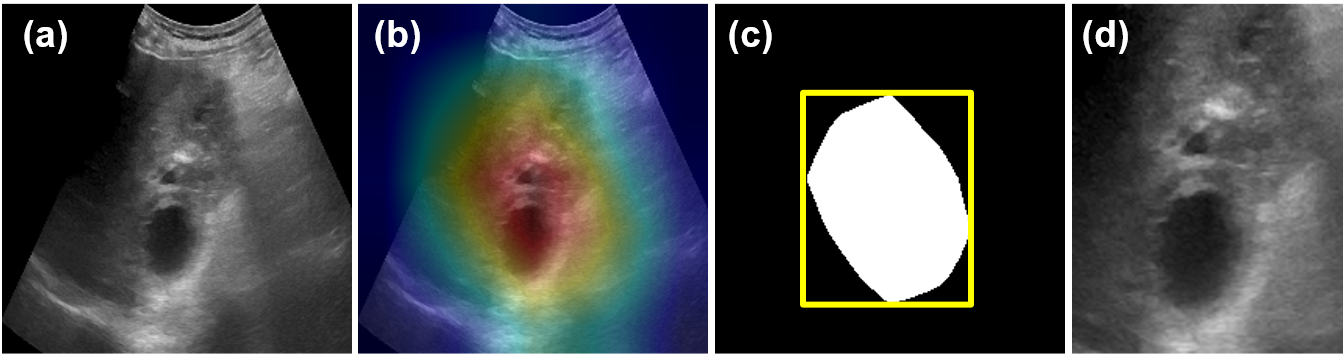}
    \caption{Generation of the ROI and the local patches from the images using the global features. (a) The original image which is the input to the global branch. (b) Activation heatmap $H$ of the features. (c) The binarized heatmap $H_B$ and the bounding box spanning it. (d) The cropped local patch to be used as the input to the local branch.}
    \label{fig:roi_mask}
\end{figure}

\section{Methods}

\subsection{Model architecture}

In general, USG is a much more challenging modality as compared to X-ray or CT scan due to the handheld nature of the sensor along with sensor noise and other artifacts. Some of the artifacts, such as shadows, often present similar visual traits of a non-malignant gallbladder lumen region \citep{basu2022surpassing}. Spurious echogenic textures from nearby organ tissues could also resemble the texture of GB mass. To tackle the above challenges, we have designed \myarch as a dual global-local attention-based model. \cref{fig:arch} shows the diagram of the proposed \myarch model. The global branch is responsible for localizing the region of interest (ROI) around the gallbladder at the whole image level. Localizing the gallbladder region helps the model mitigate effects of spurious shadows and patches that demonstrate similar visual characteristics of a gallbladder. Additionally, the global branch features help retain the necessary contextual information around the gallbladder. The ROI generated by the global branch is used to crop the local region from the entire image, and is used as input to the local branch. The local branch identifies fine-grained local features by using the bag-of-features (BOF) encoder on the local inputs. We finally use a transformer-based classification branch to fuse the global and local features and reach a superior performance. The dual attention architecture enables us with both coarse and fine-grained interpretation of GBC from USG images. The global ROI mask provides a coarse-grain explanation and tells us where to look for the cancer. The bag-of-features objectively provides the detailed, fine-grained interpretation of the malignant features of the gallbladder and helps the radiologists to explain the network decisions.

\subsubsection{Extracting global features using the global branch}

We have used a backbone inspired by ResNet-50 \citep{resnet} for extracting deep global features in the global branch. The backbone network contains 16 residual blocks, each containing $3\times3$ convolutions with residual connections and ReLU activations. When applied on the entire image, the output of the final residual block gives the global feature tensor $\mathcal{F}^g \in \mathbb{R}^{h \times w \times d}$, where $h, w$, and $d$ are the spatial height, width, and the number of channels, respectively. During the training, we formulate the global branch as a classification network for gallbladder pathology. The backbone is followed by a global average pooling, a fully connected layer with 3 output logits (as our dataset contains 3-classes, normal, benign, and malignant), and finally, a softmax layer to normalize the output of the fully connected layer. 

\subsubsection{Generating ROI from the global features} 
We use the global feature map to find the salient region in a USG image. The salient region is identified by the activation heatmap generated from the global features. Suppose $\mathcal{F}^g_k(x,y)$ denotes the activation at the $(x,y)$ spatial location of the $k$-th channel, where $x=\{1,\ldots, h\},~ y=\{1,\ldots, w\}$, and $k=\{1,\ldots, d\}$. The activations are first normalized to the range $[0,1]$. The activation heatmap, $H$, is then generated by summing the normalized activations channel-wise at each spatial location. 
\begin{align}
    H(x,y) = \sum_k {\mathcal{\widetilde{F}}^g_k(x,y)}
\end{align}
Here $\mathcal{\widetilde{F}}^g_k(x,y)$ is the normalized activation at the $(x,y)$ location of the $k$-th channel. $H(x,y)$ indicates the importance of spatial location $(x,y)$ in classifying the gallbladder abnormality, and helps identify the region of interest to focus upon. 

\par Next, we binarize the activation heatmap $H$, based on a threshold $\tau$ to identify the high activation regions. The binarized heatmap $H_B$ is obtained in the following manner, 
\begin{align}
    H_B(x,y) = \begin{cases} 1,~~ H(x,y) \geq \tau \\ 0,~~ \text{otherwise}. \end{cases}
\end{align}
It is worthy of noting that the value of $\tau$ directly controls the activation area in the binarized heatmap. A larger value of $\tau$ results in a smaller activation area, and a smaller value of $\tau$ leads to a larger activation area. We use the Otsu binarization \citep{otsu1979threshold} algorithm to select the value of $\tau$ dynamically. Apart from the Otsu binarization, we also experimented with fixed threshold values and hysteresis-based thresholding \citep{canny1986computational}. However, we found that Otsu binarization leads to superior ROI localization. We then find the bounding box $(x_{min}, y_{min}, x_{max}, y_{max})$ to cover the non-zero region in the binarized heatmap. The bounding box is used to crop this local region from the original input USG image. \cref{fig:roi_mask} shows a pictorial depiction of the entire process of finding the local region from the global features. This local region is used as input to the local branch for computing bag-of-features with local attention. 

\subsubsection{Computing local bag-of-features}

The complex dependencies between the input and the activations limit the interpretability of the DNNs. Bag-of-features (BOF) technique provides insights regarding which parts of an image are used by the networks to reach the image-level decision. The technique uses a vocabulary of visual words to represent the clusters of local patches. We adopted the BagNets-33 \citep{bagnets} architecture for using as a backbone to extract the local bag-of-features representations, $\mathcal{F}^l \in \mathbb{R}^{h'\times w' \times d}$. A single visual `word' corresponds to a channel in $\mathcal{F}^l$. Thus, we extract $d$ visual words for each image. The size of $\mathcal{F}^l$ in our experimental setup was $24\times 24\times 2048$ where $2048$ is the number of channels. The backbone network resembles ResNet, and has a convolution layer, followed by 4-layers of residual blocks, global average pooling, fully connected layer, and a softmax layer. The residual layers contain 3, 4, 6, and 3 residual blocks, respectively. The residual block layers are identical to the ResNet, except for one key difference. The first residual block of each layer employs a $3\times3$ convolution. However, differing from the ResNet architecture, the remaining residual blocks of each layer employs $1\times1$ convolutions. Instead of the standard $7\times7$ convolution followed by max-pooling used in the first layer of ResNet, the first layer of BagNets uses a $1\times1$ convolution, followed by $3\times3$ convolution. All convolution operations follow batch normalization and ReLU. The receptive field of such a local network is fixed to $33\times33$ pixels. Similar to the global branch, we train the local branch as a classification network. The local features $\mathcal{F}^l$ are aggregated using global average pooling and then passed to a linear classifier comprised of a single fully connected layer with 3-output classes. The local bag-of-features is used for finding the activations of the visual words at inference time to explain the network decisions. 
Suppose $CL_x$ denotes a convolution layer containing a $x\times x$ convolution, batch normalization, and ReLU. Further, $BN(.)$ \text{~and~} $\delta(.)$ refer to batch normalization and ReLU respectively. Each residual block $\mathcal{B}_i(.)$ is obtained as follows:
\begin{align}
    \mathcal{B}_i(\mathbf{x}) = CL_1(CL_1(CL_3(\mathbf{x})))
\end{align}
, where $\mathbf{x} \in \mathbb{R}^{h\times w \times d}$ is an input. Further,
\begin{align}
    CL_x = \delta(BN(Conv_{x \times x}))
\end{align}
, where $Conv_{x\times x}$ denotes a $x\times x$ convolution. Also, $\mathcal{F}^l$ is obtained as follows:
\begin{align}
    \mathcal{F}^l = \mathcal{B}_{16}(\ldots (\mathcal{B}_1(\mathbf{x})))
\end{align}

\subsubsection{Transformer-based feature fusion}

We use the attention mechanism of the transformer architecture \citep{vaswani2017attention, dosovitskiy2020image} to efficiently integrate global features $\mathcal{F}^g$, and local features $\mathcal{F}^l$. Transformers were originally designed for processing natural language sequences. For computational efficiency, we use global average pooling on both the feature maps and then concatenate them to form the input sequence $\mathcal{F}_0$ to the transformer. Specifically, 
\begin{align}
    \hat{\mathcal{F}^g} = GAP(\mathcal{F}^g) \in \mathbb{R}^{d} \\
    \hat{\mathcal{F}^l} = GAP(\mathcal{F}^l) \in \mathbb{R}^{d} \\
    \mathcal{F}_0 = \hat{\mathcal{F}^g} \oplus \hat{\mathcal{F}^l} \in \mathbb{R}^{2\times d}
\end{align}
where $GAP(.)$ and $\oplus$ represents global average pooling and concatenation operations, respectively. We could also directly reshape the features to form sequences of $l \times d$ where $l=h \times w$, but due to the quadratic complexity of the self-attention in the transformers, it would have abundantly increased the computation cost. The transformer branch uses the global-local features to generate the attention, and subsequently boost the important features for diagnosis. Intuitively each feature map in the bag-of-features embedding represents a `word' analogous to a bag-of-words model, and its global average represents presence or absence of the word in the image. Since each channel in the global average pooled embedding corresponds to a visual word in the bag-of-features embedding, it is possible to retrieve the corresponding visual words through backpropagating the gradient. Our design does not require the positional encoding passed to the transformer.  

\par The transformer in our architecture contains 4-transformer layers, and each layer contains multi-headed attention (MHA) and multi-layer perceptron (MLP) blocks. The multi-headed attention is an extension of the self-attention mechanism. The self-attention mechanism generates attention-weighted output sequences from an input sequence. Assuming $\mathcal{X}$ to be the input sequence, $\phi(.)$ to be the softmax function, we compute the self-attention, $\psi(.)$ as follows:
\begin{align}
    \begin{bmatrix}
    Q \\
    K \\
    V
    \end{bmatrix} = 
    \begin{bmatrix}
    W_q \\
    W_k \\
    W_v
    \end{bmatrix} \mathcal{X} \\
    \psi(\mathcal{X}) = \phi \Big( \frac{QK^T}{\sqrt{d}} \Big) V, 
\end{align}
where $W_q,~ W_k,$ and $W_v$ are the trainable weight parameters and $d$ is the dimension of $Q,~K,$ and $V$. $Q,~K,$ and $V$ refer to the `query', `key', and `value' components derived from the transformer input sequence $\mathcal{X}$. The multi-headed attention, $\mathcal{M}$, contains $m$ self-attention blocks or heads, and is computed for an input sequence $\mathcal{X}$ by,
\begin{align}
    \mathcal{M}(\mathcal{X}) = \psi_1(\mathcal{X}_1) \oplus \psi_2(\mathcal{X}_2) \oplus \ldots \oplus \psi_m(\mathcal{X}_m),
\end{align}
where $\mathcal{X}_j$'s are the splits of the input sequence $\mathcal{X}$.
The output feature, $\mathcal{F}_i$ from the $i$-th layer of the transformer is given by,
\begin{align}
    \tilde{\mathcal{F}}_i = LN\big[\mathcal{M}(\mathcal{F}_{i-1})+\mathcal{F}_{i-1}\big] \\
    \mathcal{F}_i = LN\big[MLP(\tilde{\mathcal{F}}_i) + \tilde{\mathcal{F}}_i \big],
\end{align}
where $LN(.)$ indicates layer normalization. The output of the final transformer layer ($\mathcal{F}_4$, in our case) is the fusion feature (denoted as, $\mathcal{F}^t$), that is passed through a fully connected classification head with 3 output classes representing the normal, benign, and malignant gallbladder. Note that, transformers can have additional class-tokens or a global average pooling followed by linear classifiers instead. As suggested by \citep{dosovitskiy2020image}, both approaches work similarly, but with different learning rates. In our design, we do not use any special class token, and instead use a fully connected classification head.

\subsection{Network training}
We train the global, local, and the transformer-based classifiers using the standard categorical cross-entropy (CE) loss. Assuming $W^g,~W^l,$ and $W^t$ as the parameters of the global, local, and transformer classification branches, respectively, we minimize the CE losses,
\begin{align}
    \mathcal{L}(W^g) = - \sum_{c=1}^C y_c \cdot \log{\hat{y}_c^g} ,\\
    \mathcal{L}(W^l) = - \sum_{c=1}^C y_c \cdot \log{\hat{y}_c^l} ,\\
    \mathcal{L}(W^t) = - \sum_{c=1}^C y_c \cdot \log{\hat{y}_c^t} ,
\end{align}
where $y_c$ is the ground-truth label of the $c$-th class, and $\hat{y}_c^g, \hat{y}_c^l, \hat{y}_c^t$ represent the predicted labels by the global, local, and transformer branches, respectively. $C$ is the total number of classes. For our dataset, $C=3$.
\par We use PyTorch \citep{pytorch} to implement our model. We use ImageNet \citep{imagenet} pretrained weights to initialize both the local and the global branches. We train the \myarch in multiple stages. We first fine-tune the global branch and then fine-tune the local branch on our GBC dataset. When a branch is fine-tuned, the remaining network weights are frozen. We initialize the transformer layers randomly using the default PyTorch settings. We finally unfreeze and train the entire network. We used an SGD optimizer with a learning rate of 0.005, momentum 0.9, and weight decay 0.0005. The learning rate is gradually decayed using a step learning rate scheduler. The learning rate scheduler reduces the learning rate by 20\% after every 5 epochs. We used a batch size of 16, and trained for 60 epochs. We used the same training hyper-parameters for all stages of training. We have also used resize, center crop, and normalization data augmentations as regularization methods to mitigate overfitting. The input sizes to both global and local branches after augmentation is 224$\times$224 pixels.

\subsection{Explainability of the network inferences}

We enable \myarch to provide both heatmap-based ROI visualization and the explanations of network decisions through the visual bag-of-features. The ROI visualization in the global branch shows localization of the areas of the image that provide the most discriminative features. Further, the relevant neural features from the local bag-of-features embedding are mapped to the human-understandable radiological lexicons to provide explanation of the network decision. The radiological lexicons are concurrent to the reporting and data system (RADS) standards introduced by the expert radiologists \citep{gb-rads-paper}. In summary, \myarch identifies the \emph{where} and \emph{what} of the pathology and provides a comprehensive interpretation. 
\par We have used an algorithmic approach to map the crucial neural features involved in a network decision to the radiological lexicons identified by radiologists. The radiologists first assigned lexicons to each of the images without accessing the network decisions. Our algorithm then maps the neural features with such lexicons. Such a mapping enables \myarch to generate visual words and their corresponding lexicons, and provide human-readable deep explanations of the network decisions. We discuss the approach in detail in the following subsections.

\begin{algorithm}[!t]
\small
	\caption{Lexicon and neural feature mapping}
	\label{algo}
	\SetAlgoLined
	\KwIn{Set of images and associated lexicon labels, $X=\{I_i,L_i\}_{i=1}^N$, where $L_i \in \mathcal{P}(\{l_1,\ldots,l_k\})$. $\mathcal{P}$ is the powerset and $l_j$ are individual lexicons.}
	\KwOut{Dictionaries $R$ and $M$ of key-value pairs, where in $R$, keys are lexicons and values are corresponding neural features. In the reverse map $M$, key: neural feature, value: corresponding lexicon.}
	1. Initialize set of lexicon labels $L=\phi$ \; 
	2. Set of all distinct lexicons $D=\phi$ \;
	3. Set of lexicons that do not appear individually in the images $C=\phi$ \;
	4. Initialize dictionaries $S=\{\}$ ,  $R=\{\}$, $M=\{\}$\;
	\For {i=$\{1 \ldots, N\}$}{
	    $F_i, W_i=$ Get features with gradient weights for $I_i$ \;
	    $T_i=$ \texttt{Find-Top-Feature-IDs}($W_i$) \;
	    $S[L_i]=S[L_i] \cup T_i$ \;
	    $L = L \cup \{L_i\}$ \;
	    $D = D \cup L_i$ \;
	}
	\ForEach {$l \in D$}{
	    \eIf{$l \in S.Keys()$}{
		    $R[l] = S[l]$ \;
    	}
		{
		    $C = C \cup \{l\}$ \;
        }
	}
	\ForEach {$x \in L$}{
	    \ForEach {$y \in L$} {
	        \If{($x\setminus y \in C$) and ($C \neq \phi$)} {
	            $R[x\setminus y] = S[x] \setminus S[y]$ \;
	            $C = C\setminus \{x\setminus y\}$ \;
	        }  
	    }
	} 
	\ForEach{$(K, V) \in R$}{
	    \ForEach {feature, $f \in V$}{
            $M[f] = K$ \;	    
	    }
	}
	\KwRet R, M \;
	\SetKwFunction{FMain}{Find-Top-Feature-IDs}
    \SetKwProg{Fn}{Function}{:}{}
    \Fn{\FMain{$W$}}{
        $k=\text{argmax}\{W\}$ and $W^{*} = W_k$ \;
        Initialize $T = \{k\}$\;
        \For {j=$\{1,\ldots|W|\}\setminus\{k\}$} {
            \If{$|W^{*} - W_j| \leq \epsilon$} {
                $T=T\cup j$ \;
            }
        }
        \KwRet T\;
    }
\end{algorithm}

\subsubsection{Annotating the lexicons for images}
The expert radiologists assigned the lexicons concurrent to GB-RADS for each image without accessing to the network decision. The lexicons associated with malignancy are - \emph{loss of interface with liver} and \emph{extramural invasion}. Extramural invasion can be further divided into - \emph{vascular invasion, biliary invasion}, and \emph{extramural mass}. On the other hand, \emph{intramural cyst, intramural foci}, and \emph{mural stratification} are the lexicons corresponding to benign pathologies. We denote the set of lexicons associated to malignancy as $M_L$ and the set of lexicons associated to benign pathology as $B_L$. Note that, the lexicons associated to each image sample either belongs to $M_L$, or $B_L$, or neither of them (for normal gallbladders without any pathologies). To mitigate any observer bias, we filtered only the images where the ground-truth image labels match with the image labels assigned by the radiologists. Recall that the ground-truth image labels are biopsy-proven. 

\begin{figure}[t]
    \centering
    \includegraphics[width=\linewidth]{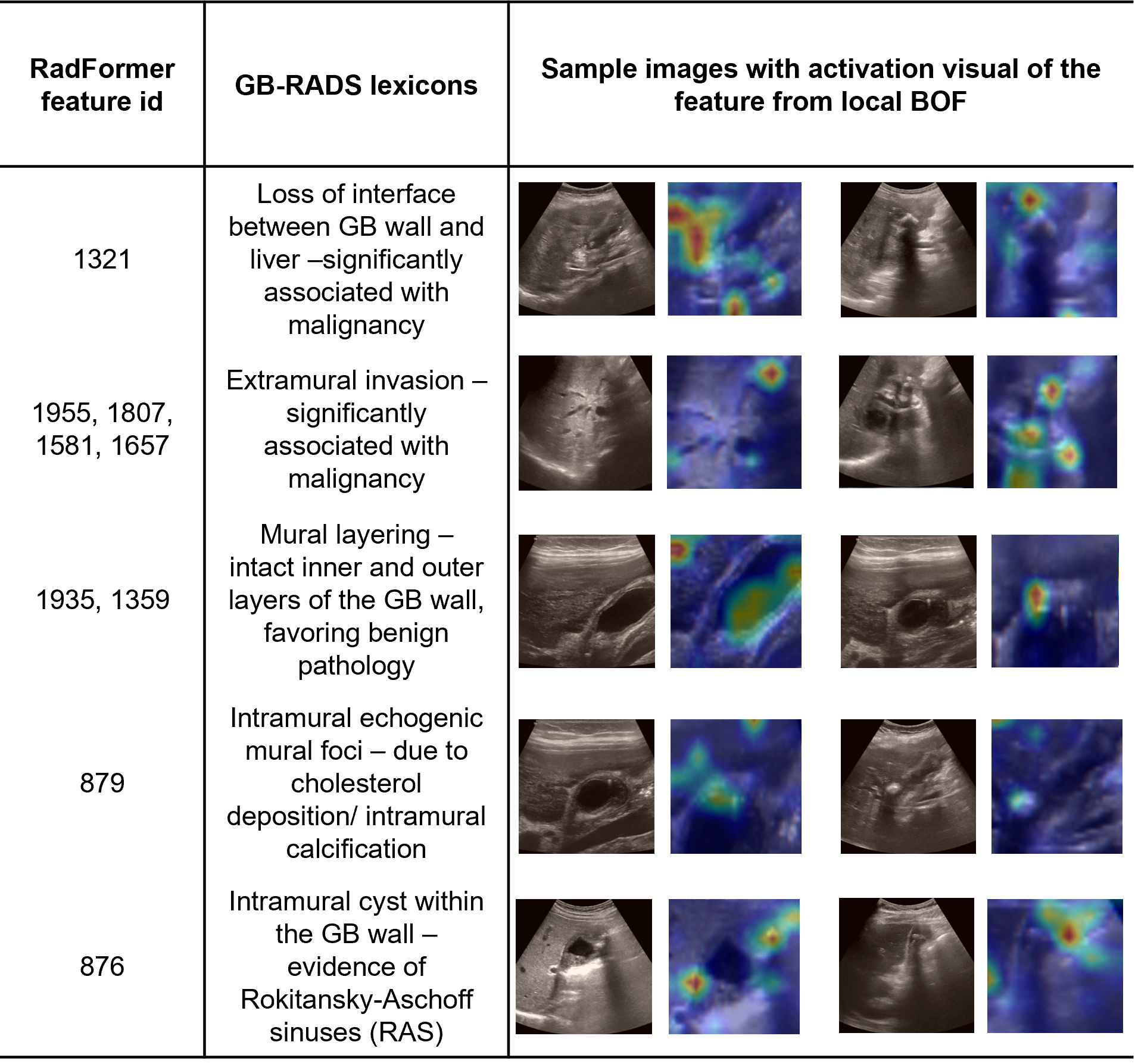}
    \caption{Human-readable lexicons identified by \myarch for detecting gallbladder pathology. Even though \myarch was trained on the image labels - normal, benign, and malignant - the local bag-of-features of the network were able to identify the fine-grained critical visual lexicons which form human-understandable explanations of the decisions. \myarch features are also well corresponded with GB-RADS lexicons introduced by \cite{gb-rads-paper}. }
    \label{fig:rads_features}
\end{figure}

\subsubsection{Mapping of lexicons and critical neural features}

We discuss the algorithm to map the lexicons and neural features in \cref{algo}. Suppose we have $k$ distinct lexicons, $\{l_1, \ldots, l_k\}$. Each image sample ($I_i$) has a lexicon label $L_i$, where $L_i$ could be a single lexicon $l_i$ or a set of multiple lexicons $l_i, \ldots, l_j$. For each image, we select the feature id with the highest gradient as the top activated feature. We also choose the feature ids whose gradients are $\epsilon-$close to the gradient of the top activated feature. These features are added to the set of neural features corresponding to the lexicon label $L_i$. We iterate over all images having the lexicon label $L_i$ to build the set of neural features associated to $L_i$. The feature ids along with their gradients for each image are extracted from the last convolution layer of the local bag-of-feature network during the inference time using backpropagation. Some lexicons (such as extramural invasions) are present in conjunction with other lexicons for the images. For such cases, we use the set difference operations to find the corresponding neural features for individual lexicons. For example, hypothetically assume we have three lexicons A, B, and C. The images present either lexicon \{A\}, or lexicons \{B, C\}, or lexicons \{A, C\}. We can find the neural feature sets corresponding to \{A\} and \{A, C\}. Then using the set difference, we can find neural features corresponding to \{C\}. Further, using the feature set of \{C\} we can decompose the feature set of \{B, C\} and find the set of features mapped to \{B\}. It's interesting to note that in the worst case, the run-time is exponential when $L$ in \cref{algo} equals the power set $\mathcal{P}$ of $\{l_1, \ldots, l_k\}$. However, the number of radiological lexicons used is very small in our case ($k=7$), and thus the exponential worst-case time complexity is not a concerning issue.

\par After the lexicon and neural feature mapping is computed using \cref{algo}, we use it during the inference to generate explanations. At inference time, for each image, we first extract the top activated feature ids as discussed in the above paragraph, and then use the feature-lexicon mapping to find the radiological lexicons corresponding to these feature ids. Additionally, the activation heatmaps of these features from the bag-of-features embedding is generated for visual interpretation. We have also asked the expert radiologists to scrutinize if the semantic meaning of the visual interpretations of the neural features were preserved. 

\par \cref{fig:rads_features} shows the mapping of lexicons to neural features for different types of gallbladders. These semantically meaningful visual features make the decision of the network transparent and retraceable. In summary, for each prediction, \myarch finds the critical features and highlights the corresponding radiological lexicons to provide a comprehensive human-understandable explanation of a decision. 

\begin{table*}[!t]
	\centering
	\resizebox{ \linewidth}{!}{%
	\begin{tabular}{llccccc}
		\toprule
		{\textbf{Group}} & {\textbf{Method}} & {\textbf{Acc.}} &  {\textbf{Spec.}} & {\textbf{Sens.}} & {\textbf{AUC}} &{\textbf{Time/image (ms)}} \\
		\midrule
		\multirow{3}{*}{CNN Classifier} &
		VGG16 \citep{vgg} & 0.693$\pm$0.036 & 0.960$\pm$0.046 & 0.495$\pm$0.234 & 0.830$\pm$0.035 & 38.98 \\ 
		& ResNet50 \citep{resnet} & 0.811$\pm$0.031 & 0.926$\pm$0.069 & 0.672$\pm$0.147 & 0.916$\pm$0.021 & 21.54 \\
		& InceptionV3 \citep{inception} & 0.844$\pm$0.039 & 0.953$\pm$0.029 & 0.807$\pm$0.097 & 0.936$\pm$0.024 & 74.58\\
		\midrule
		\multirow{3}{*}{CNN Object Detector} &
		Faster-RCNN \citep{fasterrcnn} & 0.757$\pm$0.053 & 0.840$\pm$0.046 & 0.808$\pm$0.104 & 0.903$\pm$0.047 & 96.03 \\
		& RetinaNet \citep{retinanet} & 0.749$\pm$0.073 & 0.867$\pm$0.078 & 0.791$\pm$0.089 & 0.913$\pm$0.066 & 57.21 \\
		& EfficientDet \citep{efficientdet} & 0.739$\pm$0.084 & 0.881$\pm$0.099 & 0.858$\pm$0.061 & 0.916$\pm$0.041 & 238.65  \\
		\midrule
		\multirow{3}{*}{Transformer} &
		ViT \citep{dosovitskiy2020image} & 0.803$\pm$0.078 & 0.901$\pm$0.050 & 0.860$\pm$0.068 & 0.916$\pm$0.034 & 24.62 \\
		& DEIT \citep{touvron2021training} &  0.829$\pm$0.030  & 0.900$\pm$0.040 & 0.875$\pm$0.063 & 0.933$\pm$0.021 & 23.52 \\
		& PVTv2 \citep{wang2021pvtv2} & 0.824$\pm$0.033 & 0.887$\pm$0.057 & 0.894$\pm$0.076 & 0.934$\pm$0.019 & 32.77 \\
		\midrule
		\multirow{3}{*}{Global-local} &
		GFNet \citep{wang2020glance} & 0.835$\pm$0.054 & 0.961$\pm$0.036 & 0.793$\pm$0.103 & 0.926$\pm$0.030 & 151.55 \\
		& GLiT \citep{chen2021glit} & 0.791$\pm$0.034 & 0.967$\pm$0.027 & 0.674$\pm$0.194 & 0.916$\pm$0.024 & 30.87 \\
		\midrule
		\multirow{5}{*}{Medical Imaging} &
		SRC-MT \citep{liu2020semi} & 0.825$\pm$0.035 & 0.963$\pm$0.028 & 0.776$\pm$0.105 & 0.914$\pm$0.029 & 11.23 \\
		& LibAUC \citep{yuan2021robust} & 0.815$\pm$0.028 & \textbf{0.971$\pm$0.029} & 0.701$\pm$0.152 & 0.911$\pm$0.017 & 34.39 \\
		& USCL \citep{chen2021uscl} & 0.784$\pm$0.044 & 0.895$\pm$0.054 & 0.869$\pm$0.097 & 0.878$\pm$0.046 & 27.95 \\
		& GBCNet \citep{basu2022surpassing} & 0.921$\pm$0.029 & 0.967$\pm$0.023 & 0.919$\pm$0.063 & \textbf{0.979$\pm$0.016} & 119.78 \\
		\midrule
		& \myarch (Ours) & \textbf{0.921$\pm$0.062} & 0.961$\pm$0.049 & \textbf{0.923$\pm$0.062} & 0.971$\pm$0.028 & 39.89 \\
		\bottomrule
	\end{tabular}
	}
	\caption{The 10-fold cross validation (Mean$\pm$SD) accuracy, specificity, sensitivity, and AUC (of ROC) of baselines and \myarch in detecting GBC from USG images. We have also added the average inference time per image in milli-seconds (when run in a 32GB V100 GPU) to compare the speed of detection.}
	\label{tab:perf_dnns}
\end{table*}

\begin{figure*}[t]
    \centering
    \includegraphics[width=\textwidth]{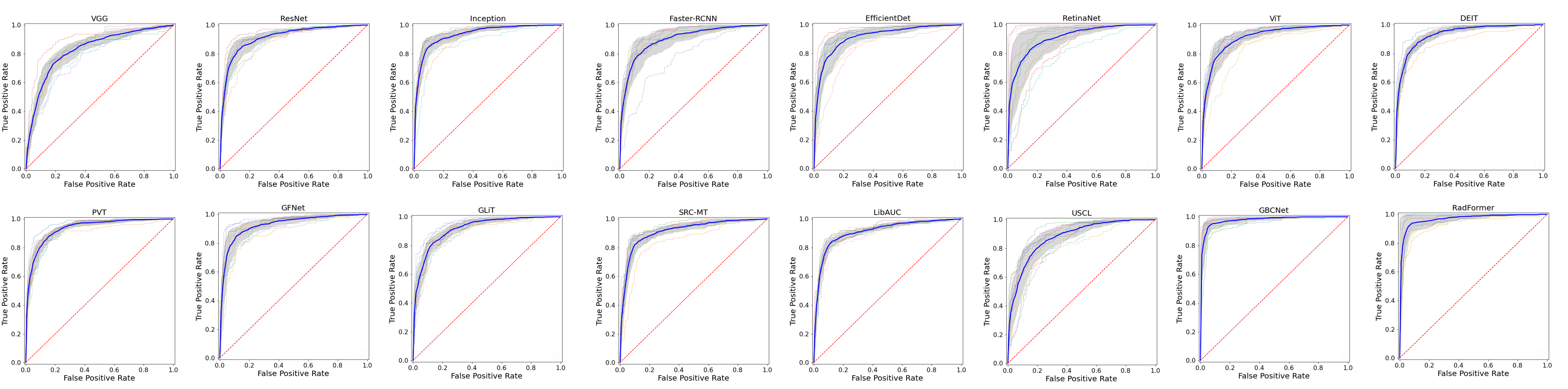}
    \caption{ROC curves for different models in classifying gallbladder malignancy. The colored lines represent curves from different cross-validation split. The dark blue curves represent the average, and the gray regions represent the standard deviation of the ROC.}
    \label{fig:roc_kfold}
\end{figure*}
\begin{figure*}
    \centering
    \begin{subfigure}[b]{0.9\textwidth}
		\centering
		\includegraphics[width=\textwidth]{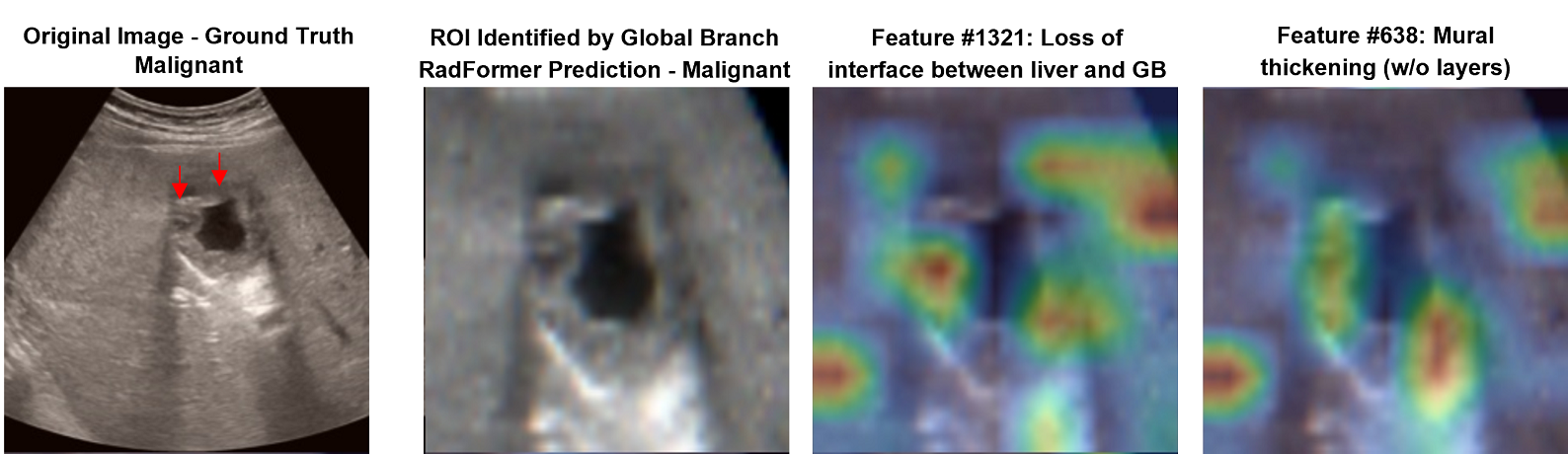}
		\caption{ Radiologists identified that the gallbladder in this image is likely to be having malignancy with a probability of 50--90\% (GB-RADS Score 4) based on the loss of interface between the liver and the gallbladder. The biopsy result confirmed malignancy. \myarch identifies the gallbladder as malignant and also correctly shows the critical features, including the loss of interface between the gallbladder and liver (feature id: 1321) and the wall thickening without layers (feature id: 638) to provide a precise and human-understandable explanation. }
	\end{subfigure}
    \begin{subfigure}[b]{0.9\textwidth}
		\centering
		\includegraphics[width=\textwidth]{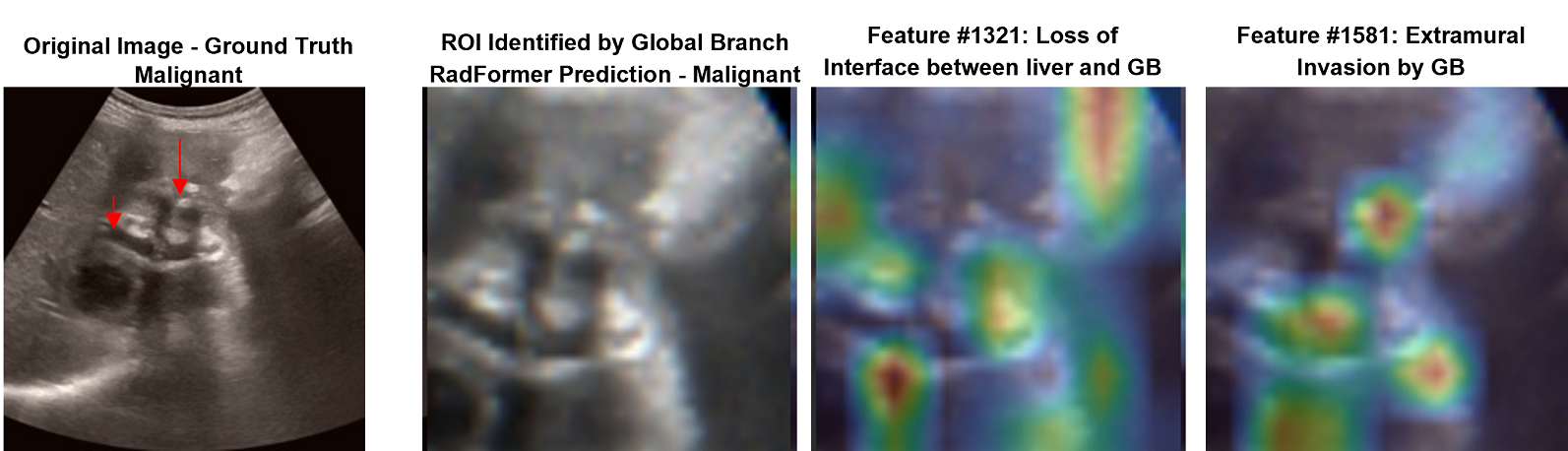}
		\caption{The gallbladder is a malignant one and shows a rare case of portal vein invasion by the malignant gallbladder.  \myarch accurately finds the feature (id:1581, extramural invasion) alongside correctly classifying the gallbladder as malignant.}
	\end{subfigure}
    \begin{subfigure}[b]{0.9\textwidth}
		\centering
		\includegraphics[width=\textwidth]{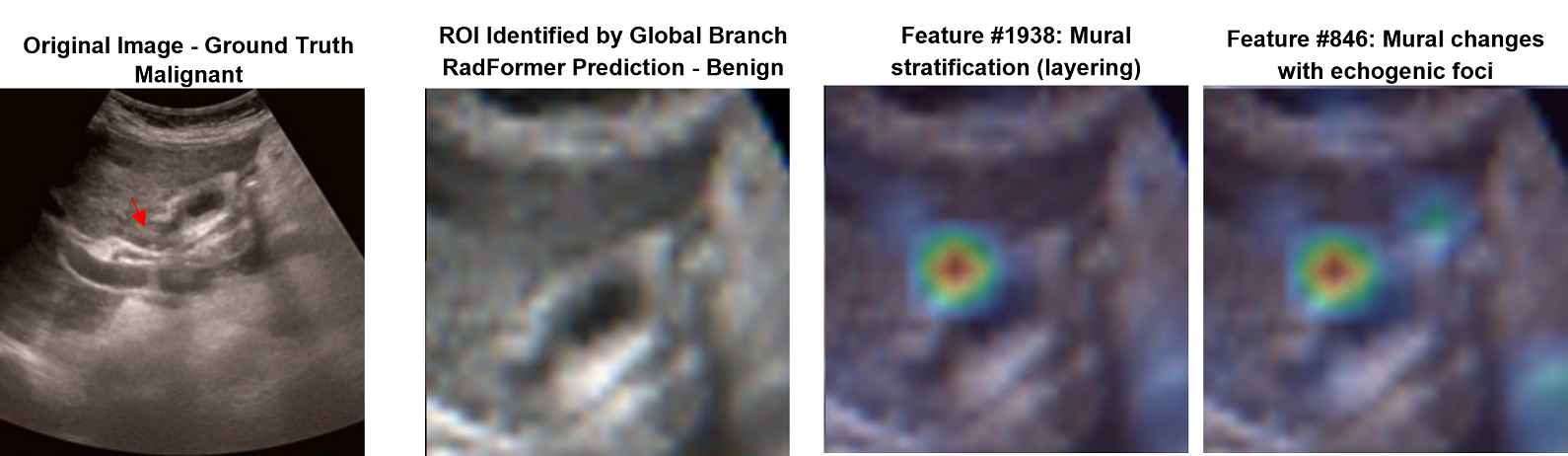}
		\caption{This is an example of \myarch missing the critical features of the malignancy and mispredicting the gallbladder as benign. There is a loss of interface between the liver and the gallbladder neck. However, \myarch only detects mural layering and the intramural echogenic focus, which are features of a benign gallbladder and miss the indistinct interface between the liver and gallbladder.}
	\end{subfigure}
	\begin{subfigure}[b]{0.9\textwidth}
		\centering
		\includegraphics[width=\textwidth]{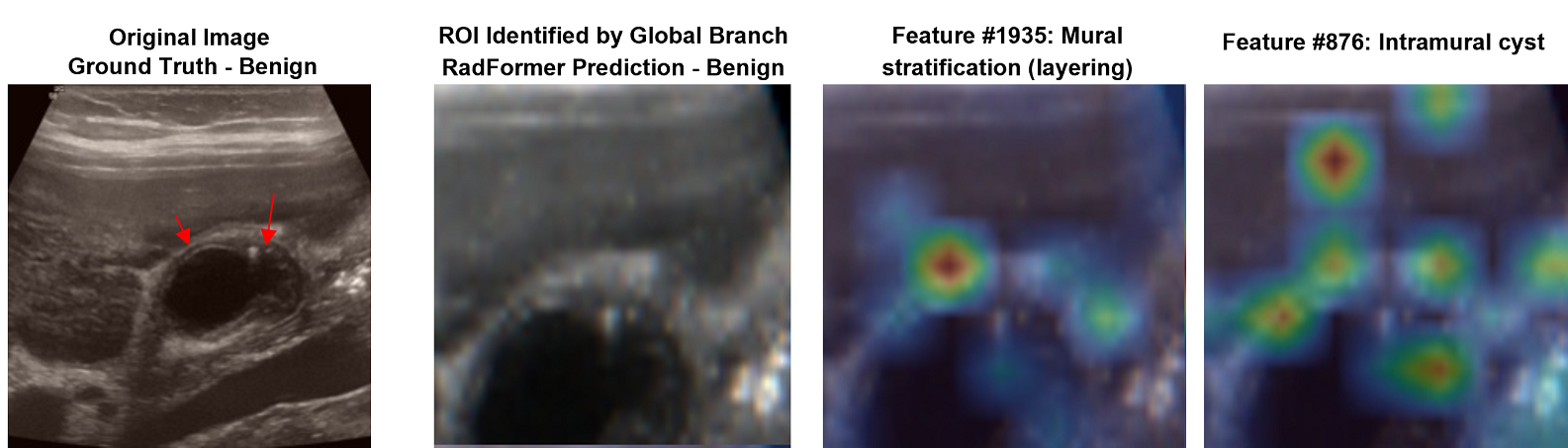}
		\caption{The gallbladder did not show signs of malignancy during the biopsy. The radiologists also classified the gallbladder as benign due to the layered appearance of the wall and the presence of intramural cyst (GB-RADS Score 2). \myarch correctly finds the features corresponding to the layered wall and cyst on the wall and successfully identifies the gallbladder as benign.}
	\end{subfigure}
    \caption{Sample explanation of the network decisions using the most critical visual features identified by \myarch. The original image with radiologists' annotation (red arrows) is on left. ROI generated by \myarch global branch, and top features from local bag-of=feature branch follows.}
    \label{fig:radform_expl}
\end{figure*}

\begin{table}[t]
	\centering
	\begin{tabular}{lccc}
		\toprule
		\textbf{Method}	& \textbf{Acc.} & \textbf{Spec.} & \textbf{Sens.} \\
		\midrule
		Radiologist A & 0.700 & 0.873 & 0.707  \\
		Radiologist B & 0.683 & 0.811 & 0.732  \\
		\midrule
		\myarch & 0.902 & 0.900 & 0.929 \\
		\bottomrule
	\end{tabular}
	\caption{We asked expert radiologists to classify malignancy for the test split of GBCU dataset \citep{basu2022surpassing} containing 122 images (31 normal, 49 benign, and 42 malignant). The ground-truth labels of these images were biopsy-proven. Radiologists were blinded from any patient data and were asked to make diagnosis based on only the USG images. \myarch significantly outperforms the human radiologists in terms of the accuracy, specificity, and sensitivity on the classification task using ultrasound image modality. The performance of the expert human radiologists is comparable to that reported in the literature \citep{bo2019diagnostic, gupta2020evaluation}.}
	\label{tab:perf_human}
\end{table}

\section{Experiments and results}
\subsection{Performance of \myarch in predicting GBC}

To assess the performance of \myarch in terms of accuracy, specificity, sensitivity, AUC, and speed of detection (inference time/ image), we compare with multiple popular convolutional neural networks and transformer-based classification models. We have used three popular deep convolutional classifiers, namely, ResNet-50 \citep{resnet}, VGG-16 \citep{vgg}, and Inception-V3 \citep{inception} as baselines. Additionally, we use recent state-of-the-art transformer-based models such as the visual transformer or ViT \citep{dosovitskiy2020image}, the pyramid transformer or PVT \citep{wang2021pvtv2} and the data-efficient image transformer or DEIT \citep{touvron2021training}. Note that we could also use object detection techniques to detect GBC from the USG images. The bounding box annotations enable us to train the deep object detectors. For completeness, we note the performance of three object detection models - Faster-RCNN \citep{fasterrcnn}, RetinaNet \citep{retinanet}, and EfficientDet \citep{efficientdet}. The object detectors may predict more than one regions, each with a predicted label. If any of the prediction instances is malignant, then the image is classified as malignant. If all the predicted regions are normal, then we label the image as normal. For all other cases, the image is predicted to be containing benign abnormalities. We show the results of using DNN models to classify GBC in \cref{tab:perf_dnns}. The reported results clearly show that the baseline object detectors and convolutional classifiers demonstrate poor sensitivity for detecting GBC from USG images. The transformer-based classifiers, ViT, DEIT, and PVT reach excellent sensitivity and detect malignancy accurately. Additionally, we have used multiple SOTA baseline methods for medical image.  analysis However, the focus of \myarch is on the interpretability as none of the baselines provide deep explanations consistent with the radiological standards. Nevertheless, \myarch achieves a high classification performance compared to the baselines. Further, our method performs at par with the SOTA methods in terms of detection speed during the inference. Thus \myarch is expected be applicable in real-world clinical setup. \cref{fig:roc_kfold} shows the area under the ROC curve (AUC) for \myarch in classifying GBC from USG images. 
\par We have also compared the GBC detection accuracy of \myarch with the human radiologists for the test split of the GBCU dataset, and observed that \myarch outperforms the human radiologists (\cref{tab:perf_human}) on the image classification task. We have blinded the radiologists from all patient-related information such as the case history, pathological test results, etc. The radiologists made their diagnosis of an image solely based on that particular image. Radiologists were evaluated based on the ground-truth image labels which are biopsy-proven.

\begin{figure}[t]
	\centering
	\includegraphics[width=0.9\linewidth]{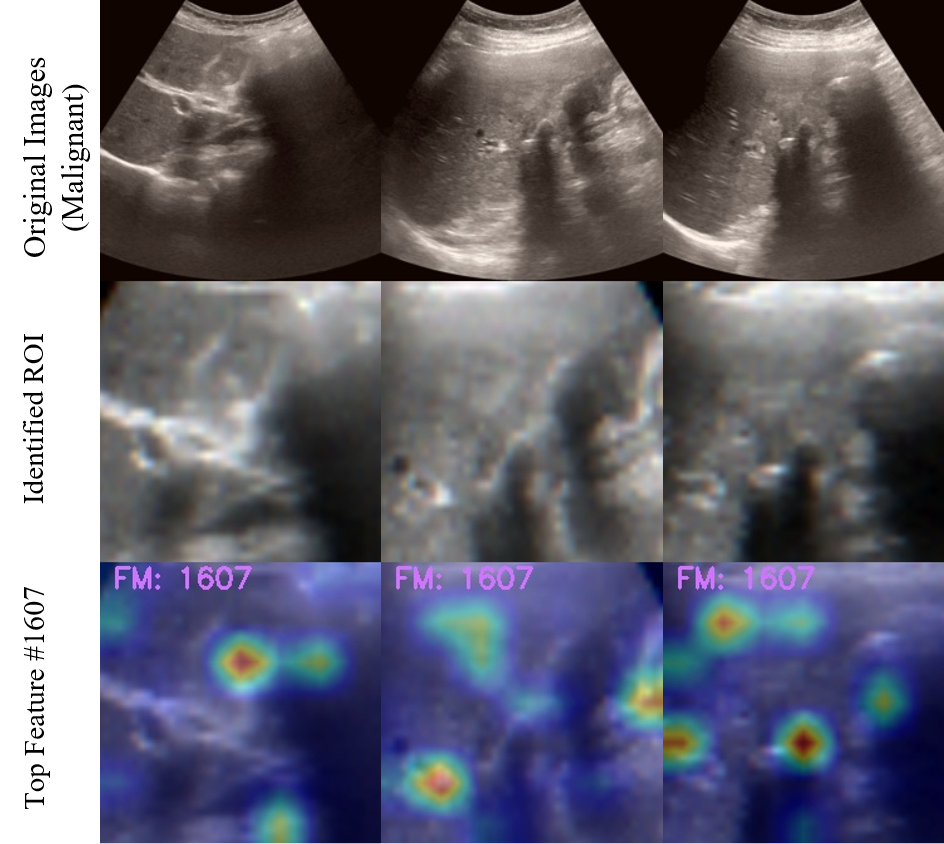}
	\caption{Feature id 1607 is within the top-$10$ activated features for 78.6\% of malignant predictions. However, radiologists could not infer the semantic meaning for this feature and suggested further clinical study to understand the precise nature of this AI-discovered feature.}
	\label{fig:unknown_feat}
\end{figure}

\subsection{Interpretability of predictions}
We show sample explanations of network decisions for a random validation set in \cref{fig:radform_expl}. The leftmost image in the figures is the original USG image with the ground-truth annotation. Next to the original image is the cropped ROI generated by the global branch for used as input to the local feature branch, and the image level prediction. The following three images on the right show the local features which are most critical for the network decision. The mapping of these local features to the lexicons form the explanation of the network decisions. The local features shown in these samples are within the top-$10$ activated features for the prediction. The \myarch explanations can identify both the ROI of the pathology and the detailed explanation that is consistent with the reporting standards used by expert radiologists. For the incorrect predictions, the explanations help identify the critical features missed by the network or the inconsistent features identified by the network. 

\subsection{Discovery of unknown malignancy indicators}

The explainable nature of \myarch also allows us to potentially discover new radiological features which may be relevant to the GBC detection. For example, we observed that the feature id 1607 was highly correlated with malignant pathologies. The feature appeared within the top-$10$ activated features for 78.6\% malignant gallbladders, indicating the feature to be suggestive of category GB-RADS 4 (malignancy is likely) in risk stratification. However, feature id 1607 was not matched with any radiological lexicon. Further, the radiologists could not infer the semantic meaning from the visual of feature 1607 as per the current medical understanding. We will require further clinical study to reveal the characterization of feature id 1607, and is a focus of our future research. If found semantically useful, we expect the GB-RADS lexicons to be updated based on such new AI discovered features. \cref{fig:unknown_feat} shows the sample activations of feature 1607 in malignant gallbladders.
A discussion on other top-activated features that were unmatched with radiological lexicons is included in Appendix D.

\begin{table*}[t]
   \centering
	\resizebox{ \linewidth}{!}{%
	\begin{tabular}{lcccccc}
		\toprule
		\textbf{Method} & \textbf{mIoU} & \textbf{mIntersection} & \textbf{ROI\_Area/Img\_Area} & \textbf{Acc.} & \textbf{Spec.} & \textbf{Sens.} \\
		\midrule
		Otsu & 0.484$\pm$0.035 & 0.934$\pm$0.016 & 0.274$\pm$0.017 & \textbf{0.921$\pm$0.062} & 0.961$\pm$0.049 & \textbf{0.923$\pm$0.062}  \\ 
		Fixed  & 0.533$\pm$0.030  & 0.851$\pm$0.028 & 0.205$\pm$0.015 & 0.917$\pm$0.065 & 0.959$\pm$0.052 & 0.911$\pm$0.057  \\ 
		Hysteresis &  0.351$\pm$0.030  & 0.987$\pm$0.008 & 0.432$\pm$0.026 & 0.918$\pm$0.064 & \textbf{0.964$\pm$0.045} & 0.896$\pm$0.092 \\
		Naive (whole image as ROI) &  0.091$\pm$0.009  & 1.000$\pm$0.000 & 1.000$\pm$0.000 & 0.844$\pm$0.032 & 0.934$\pm$0.029 & 0.810$\pm$0.062 \\
		\bottomrule
    \end{tabular}
    }
    \caption{Comparison of binarization methods in generating the ROIs from the global features. Apart from the mIoU, we assess the mean intersection of the generated ROI with the ROI annotated by radiologists. We also report the size of the predicted ROI as a fraction of image area. Recall that the bounding box annotations were not used during training the \myarch. For hysteresis-based thresholding, high and low thresholds are 120 and 50, respectively. The fixed threshold was chosen as 120. We compare the mIoU of the generated ROI with the bounding box annotations drawn by the radiologists for different binarization methods. Otsu binarization performs better in terms of accuracy and sensitivity. The high intersection value of the ROI generated by Otsu binarization indicates that the critical visual features of malignancy detection used by radiologists are well preserved in the ROI generated using Otsu binarization. The intersection is high for hysteresis-based binarization as well. However, the large ROI area indicates presence of adjacent organs and tissues, and subsequently results in lower sensitivity of malignancy detection. Similarly, naively using the entire image as ROI results in significantly lower sensitivity due to the presence of other organs.
    }
    \label{tab:perf_rois}
\end{table*}

\begin{figure}[t]
    \centering
    \includegraphics[width=\linewidth]{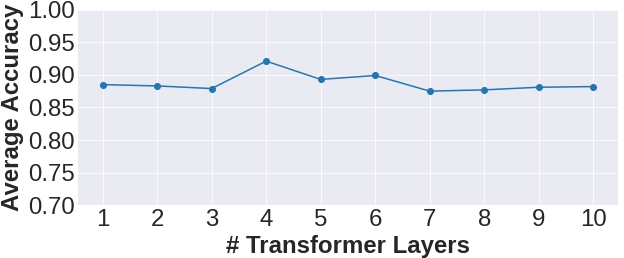}
    \caption{Variation of the 10-fold average accuracy of \myarch with the number of transformer layers. \myarch gives best accuracy with 4-transformer layers.}
    \label{fig:transformer_ablation}
\end{figure}

\begin{table}[t]
   \centering
	\resizebox{ \linewidth}{!}{%
	\begin{tabular}{llcccc}
		\toprule
		\multicolumn{2}{l}{\textbf{Method}} & \textbf{Acc.} 
		& \textbf{Spec.} & \textbf{Sens.} & \textbf{p-value (Acc.)} \\
		\midrule
		\multicolumn{2}{l}{Only global (Resnet50)} & 0.811$\pm$0.031 & 0.926$\pm$0.069 & 0.672$\pm$0.147 & $<0.01$ \\
		\midrule
		\multicolumn{2}{l}{Only local (ROI+Bagnet33)} & 0.842$\pm$0.042 & 0.941$\pm$0.058 & 0.816$\pm$0.074 & 0.022\\
		\midrule
		\multirow{3}{*}{CNN-fusion}
		& Global & 0.854$\pm$0.033 & 0.970$\pm$0.026 & 0.826$\pm$0.062 & 0.032 \\
		& Local & 0.859$\pm$0.028 & 0.943$\pm$0.047 & 0.861$\pm$0.072 & 0.034\\
		& Fusion & 0.873$\pm$0.035 & 0.970$\pm$0.033 & 0.862$\pm$0.078 & 0.044\\
		\midrule
		\multirow{3}{*}{Transformer-fusion}
		& Global & 0.878$\pm$0.036 & 0.966$\pm$0.027 & 0.865$\pm$0.088 & 0.044\\
		& Local & 0.894$\pm$0.031 & 0.952$\pm$0.034 & 0.905$\pm$0.079 & 0.047\\
		& Fusion & 0.921$\pm$0.062 & 0.961$\pm$0.049 & 0.923$\pm$0.062 & -- \\
		\bottomrule
	\end{tabular}
	}
    \caption{Ablation study related to the significance of the design components - global branch, local branch, and attention-based feature fusion through transformers. We also show the p-value of accuracy for pairwise t-test of all other ablated models with RadFormer (Fusion branch of Transformer-fusion model). The p-value indicates the performance improvement of RadFormer accuracy to be statistically significant ($p<0.05$).}
    \label{tab:ablation-parts}
\end{table}

\subsection{Ablation study}
\subsubsection{Binarization techniques for ROI generation}
The performance of \myarch and the preciseness of the fine-grained visual lexicons depend strongly on the quality of the ROI predicted by the global features. We quantitatively evaluate  the predicted ROIs with the ROI bounding boxes annotated by the radiologists. In \cref{tab:perf_rois}, we report the performance of \myarch for different binarization methods used for localizing the ROI from the global features. We found that although the intersection-over-union (IoU) of the predicted ROI with the ROI identified by radiologists (GT) are low, the intersection of the predicted ROI with GT is large (typically over 90\%). This is an indicator of the predicted ROI being able to actually retain most of the visual information used by a radiologist for GBC detection. A naive way of selecting the entire GT is to pass the entire image as the ROI. However, we found that such naive scheme results in significant performance degradation, due to the interference of other organs and tissues. In contrast, the size of the predicted ROI is only about 27\% of the entire image for Otsu binarization, which indicates that the ROI is not arbitrary, but precise in retaining the critical visual cues of GBC.

\subsubsection{Depth of transformer branch}
We experimented with varying the number of transformer layers in the \myarch network to identify the optimal number of layers. \cref{fig:transformer_ablation} shows the variation of the 10-fold average accuracy with the number of transformer layers. It can be seen that keeping the number of transformer layers to 4, leads to the optimal performance of \myarch.

\subsubsection{Efficacy of the proposed design}
We show in \cref{tab:ablation-parts} the effect of different components of the \myarch on the performance. The global path is ResNet50 and the local path is BagNets33 applied on random ROIs. The standalone global or local paths do not lead to adequate performance. The global-local architecture on the other hand, results in excellent performance boost for both pathways. Both the CNN-based and transformer-based fusion results in improvement as compared to the global or local components of the \myarch. However, the design of attention-based feature fusion using the transformer results in best sensitivity and accuracy of GBC detection.
\begin{figure}[t]
	\centering
	\includegraphics[width=0.9\linewidth]{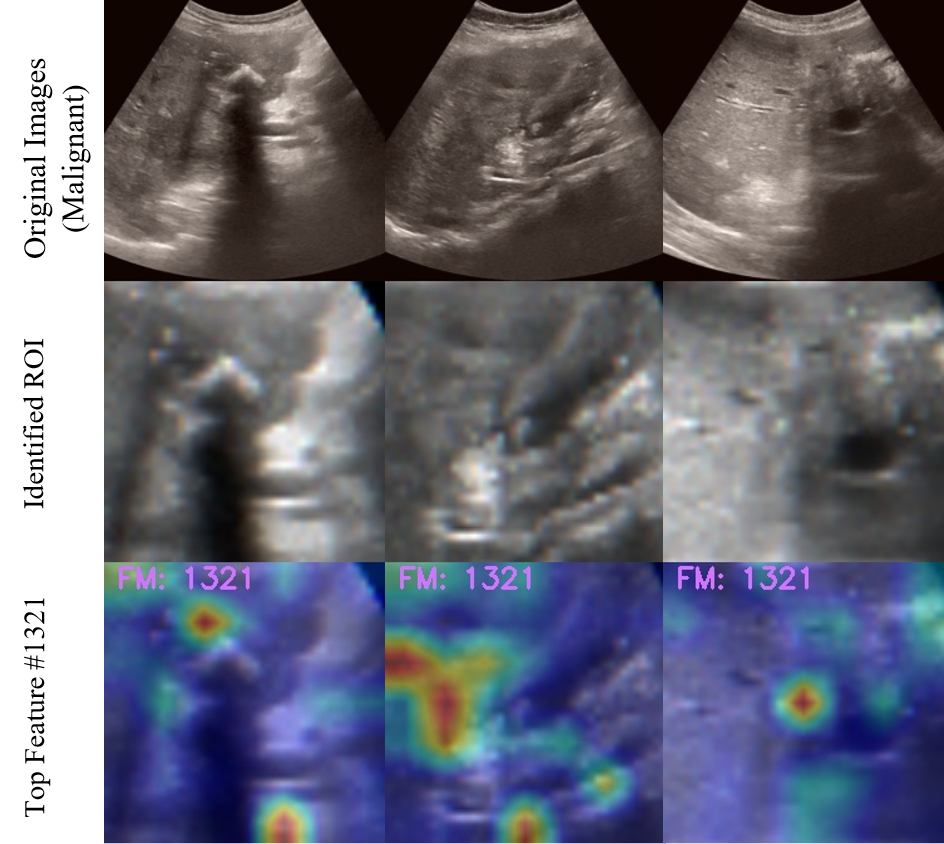}
	\caption{Feature id 1321 corresponds to the loss of interface between the liver and the gallbladder and was one of the top activated features for all these pathological gallbladders. We show some sample correspondence of feature id 1321 with the loss of interface between liver and gallbladder. 92.9\% of malignant gallbladders showed loss of interface with the liver during our assessment. This probability score is consistent with the GB-RADS probability score of $>\!90$\% (risk category 5) for malignancy if an indistinct interface between the liver and gallbladder is found.}
	\label{fig:malg_corr_1}
\end{figure}

\begin{figure}[t]
    \centering
    \includegraphics[width=0.9\linewidth]{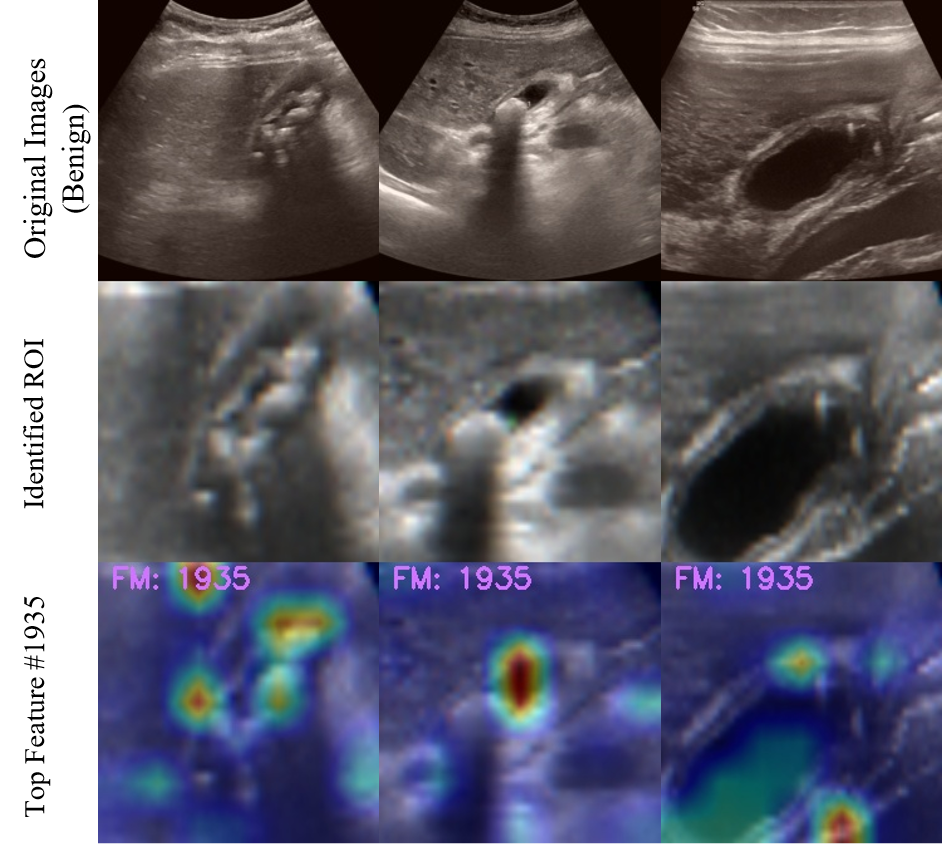}
    \caption{We assess \myarch decisions for benign gallbladders showing the intramural stratification pathology (layered appearance of the gallbladder wall). Feature id 1935 corresponds to the layered appearance of the wall and was strongly activated for all such pathological cases.}
    \label{fig:ben_corr_1}
\end{figure}
\begin{figure}[t]
    \centering
    \includegraphics[width=\linewidth]{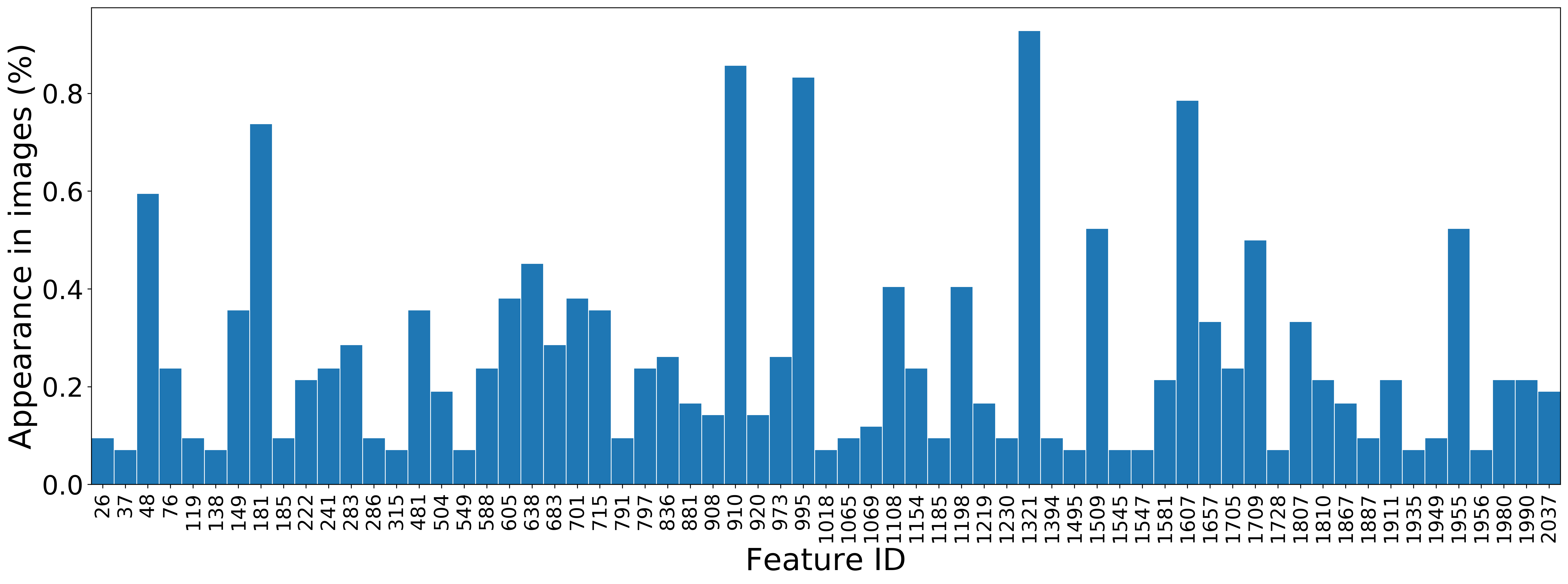}
    \caption{The frequency of the top-10 activated feature ids in the set of malignant images. Such a frequency map provides a measure of correlation of a feature id with malignancy.}
    \label{fig:correlation_freq}
\end{figure}

\section{Discussion and conclusion}
GBC is a severe affliction affecting the lives of many patients. The high mortality rate of GBC is attributed to the delay in diagnosis. GBC lacks specific clinical features to aid early detection and is usually not detected before a metastasized state. Transabdominal USG is a popular diagnostic imaging modality for visualizing the gallbladder due to its low-cost, non-radiation, and broad accessibility, making it the candidate modality for GBC risk stratification. GBC presents wall thickening, lesions, or mass \citep{lopes2021gallbladder}. However, discerning benign and malignant characteristics of wall thickening remain challenging in USG \citep{gupta2020evaluation}. Recently, \cite{gb-rads-paper} introduced GB-RADS (gallbladder reporting and data system) to identify clinical features of GBC presented in USG systematically. 
\par For a non-malignant gallbladder, the liver and gallbladder wall interface is well delineated. \emph{Loss of interface with the liver} is strongly associated with gallbladder malignancy \citep{catalano2008mr}. Similarly, a normal gallbladder consists of a thin, uniform wall. For benign abnormalities, often, the wall is uniformly thickened and demonstrates a layered appearance (\emph{mural layering}) with clear visualization of the inner and outer wall \citep{mizuguchi1997endoscopic}. \emph{Intramural cyst or echogenic foci} are also crucial identifiers of a benign pathology.
On the other hand, the presence of non-uniform, \emph{asymmetric wall thickening} is an indication of malignant abnormality. Malignant wall thickening also loses the layered appearance as the mucosa is disrupted with infiltration into the deeper layers \citep{joo2013differentiation}. 
\par We observed that despite the access to objective RADS features, identifying GBC is hard for human radiologists. Subtle signs of malignancy can be missed for cases with complicated acute cholecystitis \citep{liu2012contrast}. Early-stage (T1) GBC could show mural layering. Xanthogranulomatous cholecystitis could be inaccurately associated with a high likelihood of malignancy \citep{deng2015xanthogranulomatous, zhang2019usefulness}. \myarch, on the other hand, provides super-human accuracy of GBC detection and allows investigating the local bag of visual words embedding for explaining the network decisions consistent with the RADS standards. The feature correspondences to the pathologies are the key to building the radiologist-like interpretations of the network inferences. It is critical to validate the correlation between the fine-grained interpretable features and the image level network decision to assert the preciseness of \myarch explanations.
\par We demonstrate how the feature 1321 is correlated with the malignant pathology, and feature 1935 is correlated with the benign gallbladders. Recall that (\cref{fig:rads_features}) the lexical meaning of feature 1321 is \emph{loss of interface between the liver and gallbladder}, and feature 1935 is \emph{mural layering}. 
We give the pathological cases to \myarch for inference and check which features are activated to validate the feature correspondence. The frequency of each feature map in the set of malignant images provides a measure of correlation of a feature with malignancy. Such a frequency plot is shown in \cref{fig:correlation_freq}. \cref{fig:malg_corr_1} show the correspondence of feature id 1321 with the network decisions for malignant gallbladders. Similarly, \cref{fig:ben_corr_1} show correspondence of feature id 1935 with benign gallbladders. 
\par It is worthy to note that \myarch could be used as a second reader by inexperienced radiologists to mitigate the effect of inter-observer variance in GB-RADS. Further, calculating the probability of malignancy given the activated features would help to update the GB-RADS risk scores. We have also shown the ability of \myarch in identifying unknown features. The AI-assisted features could be characterized by clinical study and integrated into the GB-RADS in the future. Since the \myarch design methodology is non-specific to GBC, we expect it to be useful for generating explainable decisions for other diseases as well. In future, we would like to assess the applicability of \myarch on other types of cancers.

\section*{Supplementary Material}

\appendix
\beginsupplement

\section{Effect of using other backbones}
\begin{table}[t]
   \centering
	\begin{tabular}{lccc}
		\toprule
		\textbf{Backbone} & \textbf{Accuracy} 
		&  \textbf{Specificity} & \textbf{Sensitivity}  \\
		\midrule
		ResNet18 & 0.885$\pm$0.049 & 0.964$\pm$0.042 & 0.908$\pm$0.078 \\
		ResNet34 & 0.883$\pm$0.034 & 0.950$\pm$0.038 & 0.903$\pm$0.084 \\
		ResNet50 & 0.921$\pm$0.062 & 0.961$\pm$0.049 & 0.923$\pm$0.062 \\
		\bottomrule
	\end{tabular}
	\caption{Effect of the choice of global branch backbone on \myarch performance. The local branch is Bagnets-33 for all three cases.}
    \label{tab:ablation-backbone}
\end{table}
We have experimented with different variants of ResNet backbones for the global branch in the \myarch architecture. The performances are reported in \cref{tab:ablation-backbone}. \myarch with ResNet50 global backbone obtains the highest accuracy and sensitivity, while \myarch with ResNet18 backbone obtains the highest specificity.

\section{Attention score visualization in transformer}
\begin{figure}
    \centering
    \includegraphics[width=\linewidth]{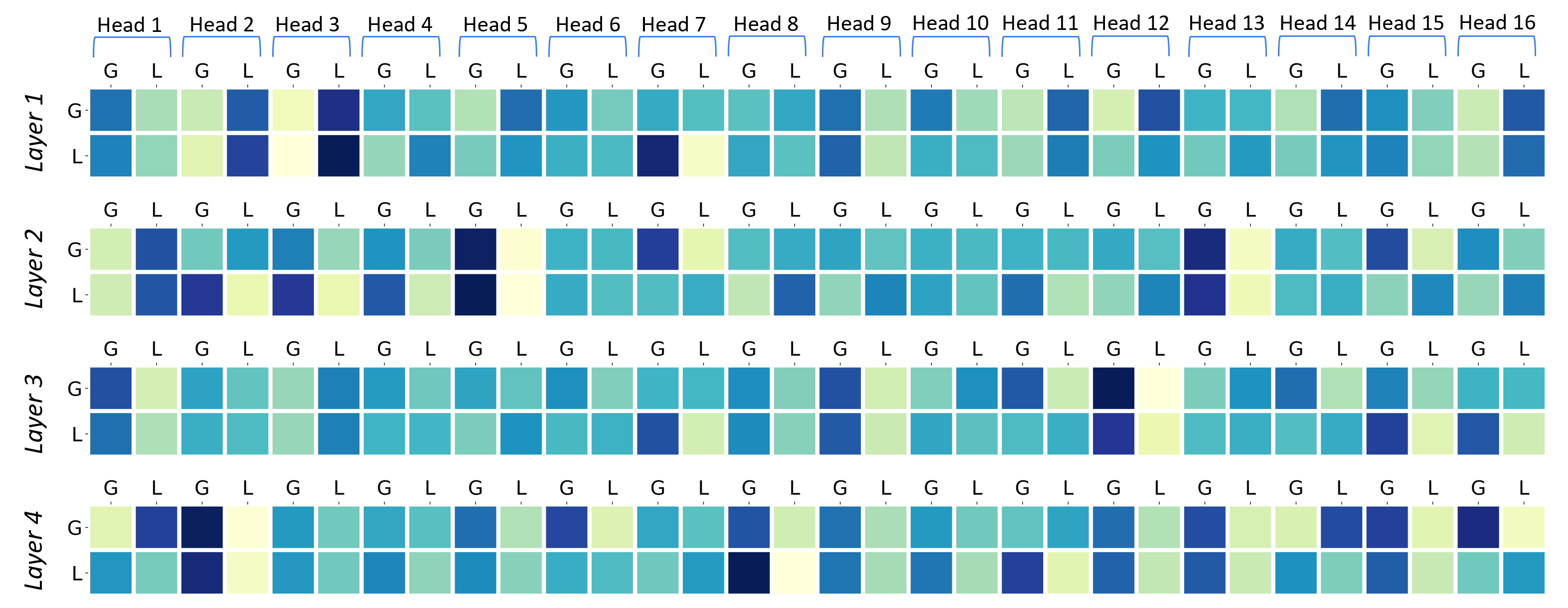}
    \caption{Heatmap of the attention scores of global and local tokens in different transformer layers across all 16 attention heads. The scores are averaged over malignant samples. Low scores are represented with lighter and high scores are represented with darker shades.}
    \label{fig:attn_score}
\end{figure}

\cref{fig:attn_score} shows the visualization of pooled attention scores in different attention heads across the layers. It can be seen that both the global and local tokens influence each other by putting attention to self and the other token. This is an indicator that the attention mechanism is efficiently fusing the global and local features.

\section{Class-wise performance of models}
\begin{table}[h]
   \centering
	\resizebox{ \linewidth}{!}{%
	\begin{tabular}{lcccc}
		\toprule
		\multirow{2}{*}{\textbf{Method}} & \multicolumn{4}{c}{\textbf{Accuracy}} \\
		&  \textbf{Overall} & \textbf{Nml.} & \textbf{Ben.} & \textbf{Malg.} \\
		\midrule
		ResNet50 & 0.811$\pm$0.031 & 0.904$\pm$0.040 & 0.812$\pm$0.051 & 0.672$\pm$0.147 \\
		Faster-RCNN & 0.757$\pm$0.053 & 0.720$\pm$0.046 & 0.767$\pm$0.061 & 0.808$\pm$0.104 \\
		ViT & 0.803$\pm$0.078 & 0.764$\pm$0.081 & 0.786$\pm$0.147 & 0.860$\pm$0.068 \\
		GFNet & 0.835$\pm$0.054 & 0.882$\pm$0.053 & 0.829$\pm$0.083 & 0.793$\pm$0.103 \\
		SRC-MT & 0.825$\pm$0.035 & 0.944$\pm$0.040 & 0.764$\pm$0.061 & 0.776$\pm$0.105 \\
		LibAUC & 0.815$\pm$0.028 & 0.886$\pm$0.067 & 0.813$\pm$0.070 & 0.701$\pm$0.152 \\
		GBCNet & 0.921$\pm$0.029 & 0.902$\pm$0.085 & 0.920$\pm$0.052 & 0.919$\pm$0.063 \\
		\midrule
		RadFormer & 0.921$\pm$0.062 & 0.949$\pm$0.069 & 0.906$\pm$0.086 & 0.923$\pm$0.062\\
		\bottomrule
	\end{tabular}
	}
    \caption{Class-wise performance of different models. We report the per-class accuracy for normal, benign, and malignant gallbladders.}
    \label{tab:classwise}
\end{table}

We have also reported the class-wise performance of \myarch and some selected baselines in \cref{tab:classwise}. \myarch works well across different classes. 

\section{Features unmatched to lexicons}
\begin{figure}
    \centering
    \includegraphics[width=\linewidth]{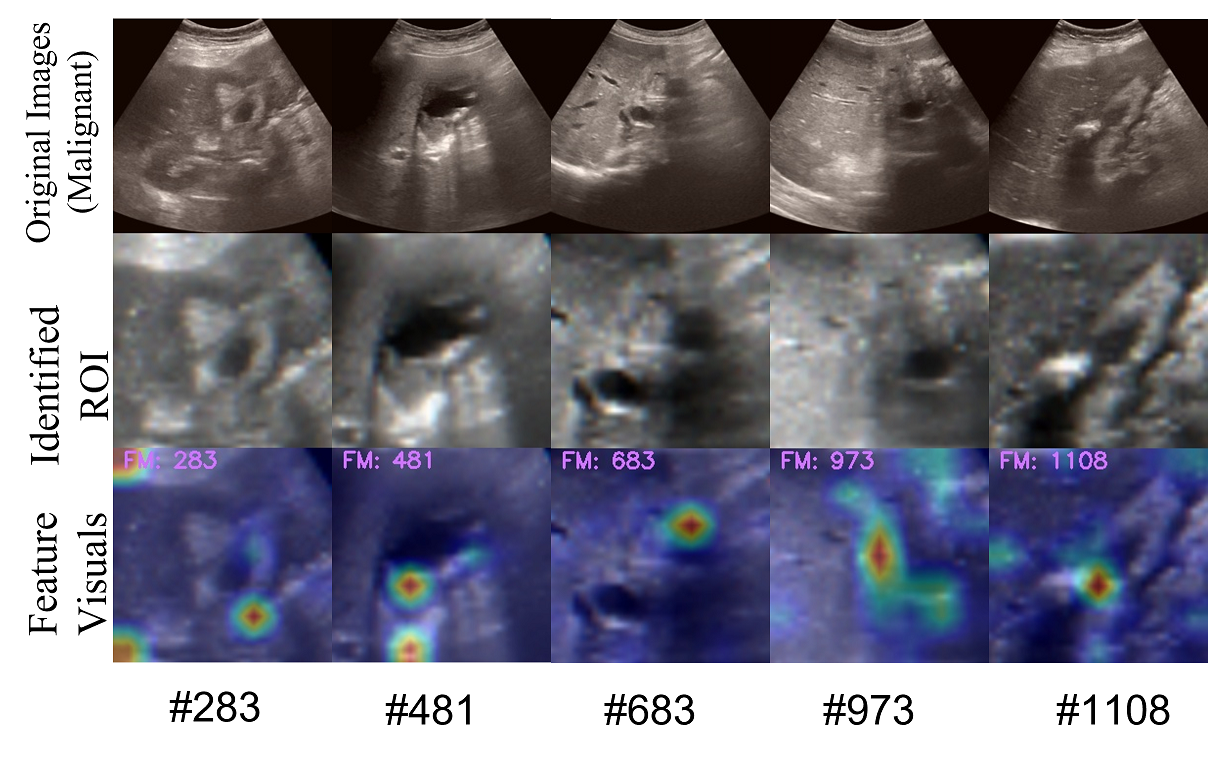}
    \caption{Visuals of a few top-actived features from the malignant samples that were not matched to any radiological lexicon by our algorithm.}
    \label{fig:unknown-new}
\end{figure}

There are a total of 65 features in the set of top-10 activated features for each of the malignant samples. Our algorithm matched 32 of the top activated features to radiologist approved lexicons identified in GB-RADS. That is, about 50\% of the set of top-10 malignant features are explainable with our algorithm. However, although there are about 33 features that are unmatched to radiological lexicons, we observed only one of them (id 1607) has been activated for 78\% malignant samples. There are 5 other unmatched features (ids: 283, 481, 683, 973, 1108) that were among the top-10 activated features for 25--35\% of malignant samples. \cref{fig:unknown-new} shows the visualization of these five features. Out of the remaining features, 19 have been activated for less than 10\% and 8 have been activated for 10--25\% malignant samples.

\bibliographystyle{model2-names.bst}\biboptions{authoryear}
\bibliography{Medima_102676}

\end{document}